\documentclass[sigconf]{acmart}

\usepackage{color}
\usepackage{colortbl}
\usepackage[nolist]{acronym}
\usepackage{graphicx}
\usepackage{amsfonts}
\usepackage{amsmath}
\usepackage{amsthm}
\usepackage{enumitem}
\usepackage{algorithmic}
\usepackage{subfig}

\usepackage[export]{adjustbox}
\usepackage{tcolorbox}
\usepackage{diagbox}
\usepackage{tabularx}
\usepackage{booktabs}                   % Nicer tables
\usepackage{multicol}                   % Content of a table over several columns
\usepackage{multirow}                   % Content of a table over several rows
\usepackage{makecell}
\usepackage{rotating}                   % Alows to rotate text and objects
\usepackage{enumitem}
\usepackage{url}
\usepackage{footnote}
\setcopyright{none}
\renewcommand\footnotetextcopyrightpermission[1]{} % removes footnote with conference information in first column
\newcommand{\et}[1]{#1~et~al.}

\AtBeginDocument{%
  \providecommand\BibTeX{{%
    \normalfont B\kern-0.5em{\scshape i\kern-0.25em b}\kern-0.8em\TeX}}}

\setcopyright{acmcopyright}
\copyrightyear{2021}
\acmYear{2021}
\acmDOI{10.xx/xxxxxxx.xxxxxxx}

\acmConference[HRI '21 LEAP-HRI Workshop]{HRI '21: Workshop on Lifelong Learning and Personalization in Long-Term Human-Robot Interaction (LEAP-HRI), ACM/IEEE International Conference on Human-Robot Interaction (HRI2021), Virtual}{March, 2021}{Virtual}
\acmBooktitle{HRI '21: Workshop on Lifelong Learning and Personalization in Long-Term Human-Robot Interaction (LEAP-HRI), HRI2021]}

\begin{document}

% \title{An Approach for Mitigating Bias in Facial Expression Recognition: Continual Learning}
\title{Towards Fair Affective Robotics: Continual Learning for Mitigating Bias in Facial Expression and Action Unit Recognition}

\author{Ozgur Kara}
\affiliation{%
      \institution{Electrical \& Electronics Engineering Department\\Bogazici University}
      \city{Istanbul}
      \country{Turkey}
}
\email{ozgur.kara@boun.edu.tr}
\orcid{0000-0003-1401-1723}

\author{Nikhil Churamani}
\orcid{0000-0001-5926-0091}

\affiliation{%
      \institution{Department of Computer Science and Technology \\University of Cambridge}
      \city{Cambridge}
      \country{United Kingdom}      
}
\email{nikhil.churamani@cl.cam.ac.uk}

\author{Hatice Gunes}
\affiliation{%
      \institution{Department of Computer Science and Technology \\University of Cambridge}
      \city{Cambridge}
      \country{United Kingdom}      
}
\email{hatice.gunes@cl.cam.ac.uk}
\orcid{0000-0003-2407-3012}
\thanks{O.~Kara contributed to this work while undertaking a summer research study at the Department of Computer Science and Technology, University of Cambridge. N.~Churamani is funded by the EPSRC under grant EP/R$513180$/$1$ (ref.~$2107412$). H.~Gunes' work is supported by the EPSRC under grant ref. EP/R$030782$/$1$. 
%and partially by the European Union's Horizon~$2020$ Research and Innovation programme under grant agreement No.~$826232$. 
The authors also thank Prof Lijun~Yin from Binghamton University (USA) for providing access to the BP4D Dataset and the relevant race attributes; and Shan~Li, Profs Weihong~Deng and JunPing~Du from Beijing University of Posts and Telecommunications (China) for providing access to RAF-DB.
}

\begin{abstract}
    As affective robots become integral in human life, these agents must be able to fairly evaluate human affective expressions without discriminating against specific demographic groups. Identifying \textit{bias} in \ac{ML} systems as a critical problem, different approaches have been proposed to mitigate such biases in the models both at data and algorithmic levels. In this work, we propose \acf{CL} as an effective strategy to enhance \textit{fairness} in \acf{FER} systems, guarding against biases arising from imbalances in data distributions. We compare different state-of-the-art bias mitigation approaches with \ac{CL}-based strategies for \textit{fairness} on expression recognition and \acf{AU} detection tasks using popular benchmarks for each; RAF-DB and BP4D. Our experiments show that \ac{CL}-based methods, on average, outperform popular bias mitigation techniques, strengthening the need for further investigation into \ac{CL} for the development of \textit{fairer} \ac{FER} algorithms.
\end{abstract}

%%
%% The code below is generated by the tool at http://dl.acm.org/ccs.cfm.
%% Please copy and paste the code instead of the example below.
%%
% \begin{CCSXML}
% <ccs2012>
%  <concept>
%   <concept_id>10010520.10010553.10010562</concept_id>
%   <concept_desc>Computer systems organization~Embedded systems</concept_desc>
%   <concept_significance>500</concept_significance>
%  </concept>
%  <concept>
%   <concept_id>10010520.10010575.10010755</concept_id>
%   <concept_desc>Computer systems organization~Redundancy</concept_desc>
%   <concept_significance>300</concept_significance>
%  </concept>
%  <concept>
%   <concept_id>10010520.10010553.10010554</concept_id>
%   <concept_desc>Computer systems organization~Robotics</concept_desc>
%   <concept_significance>100</concept_significance>
%  </concept>
%  <concept>
%   <concept_id>10003033.10003083.10003095</concept_id>
%   <concept_desc>Networks~Network reliability</concept_desc>
%   <concept_significance>100</concept_significance>
%  </concept>
% </ccs2012>
% \end{CCSXML}

% \ccsdesc[500]{Computer systems organization~Embedded systems}
% \ccsdesc[300]{Computer systems organization~Redundancy}
% \ccsdesc{Computer systems organization~Robotics}
% \ccsdesc[100]{Networks~Network reliability}

\maketitle
\thispagestyle{empty}


\begin{figure*}
    \centering
    \subfloat[Baseline CNN Architecture implementing a ResNet-based architecture.\label{fig:baseline}]{\includegraphics[width=0.6\textwidth,valign=c]{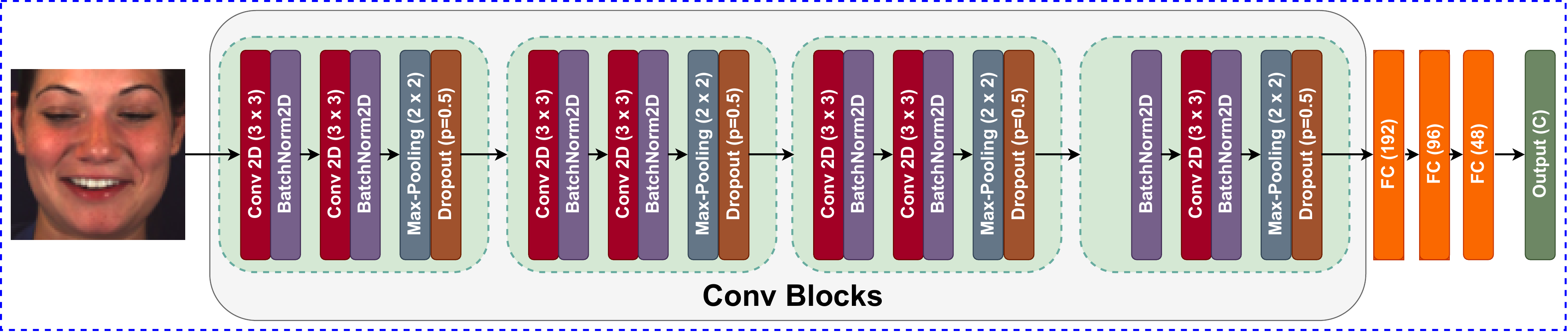}}\\%\vspace{-2mm}

    \subfloat[\acs{DDC} Architecture implementing an N$\times$M Classifier.\label{fig:ddc}]{\includegraphics[width=0.337\textwidth,valign=c]{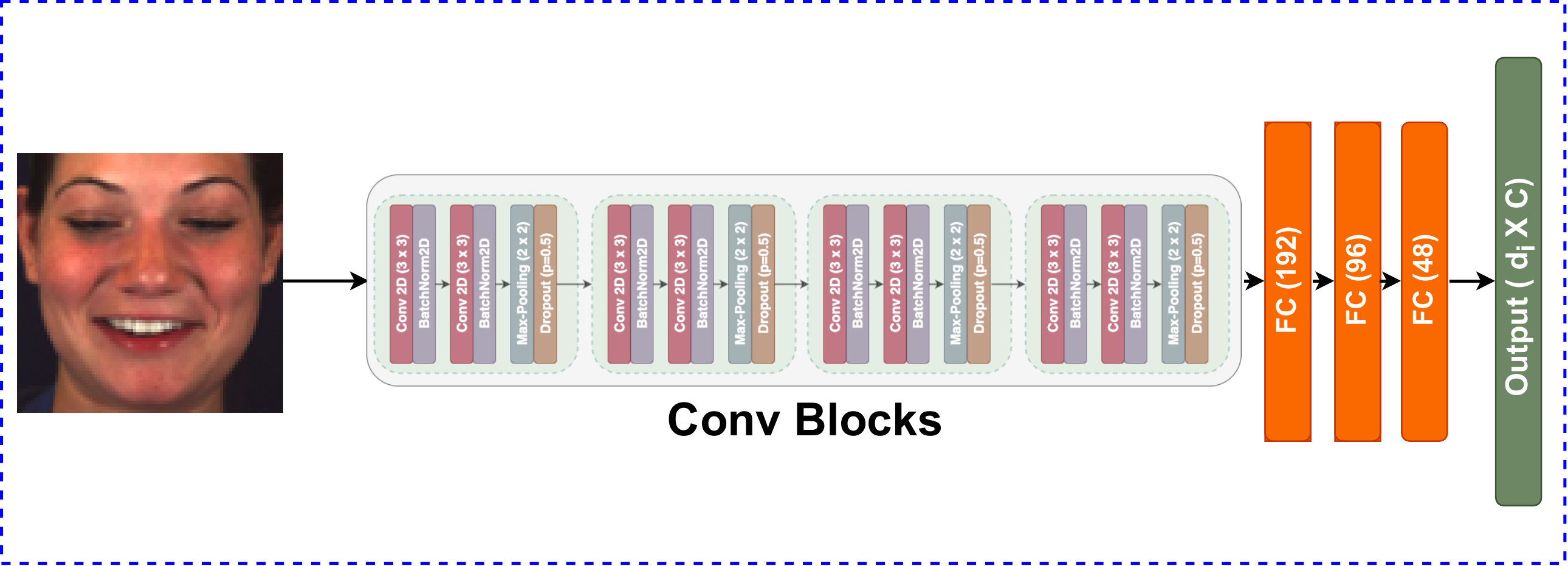}}\hspace{1em}
    \subfloat[\acs{DIC} Architecture with separate classifiers for each domain.\label{fig:dic}]{\includegraphics[width=0.337\textwidth,valign=c]{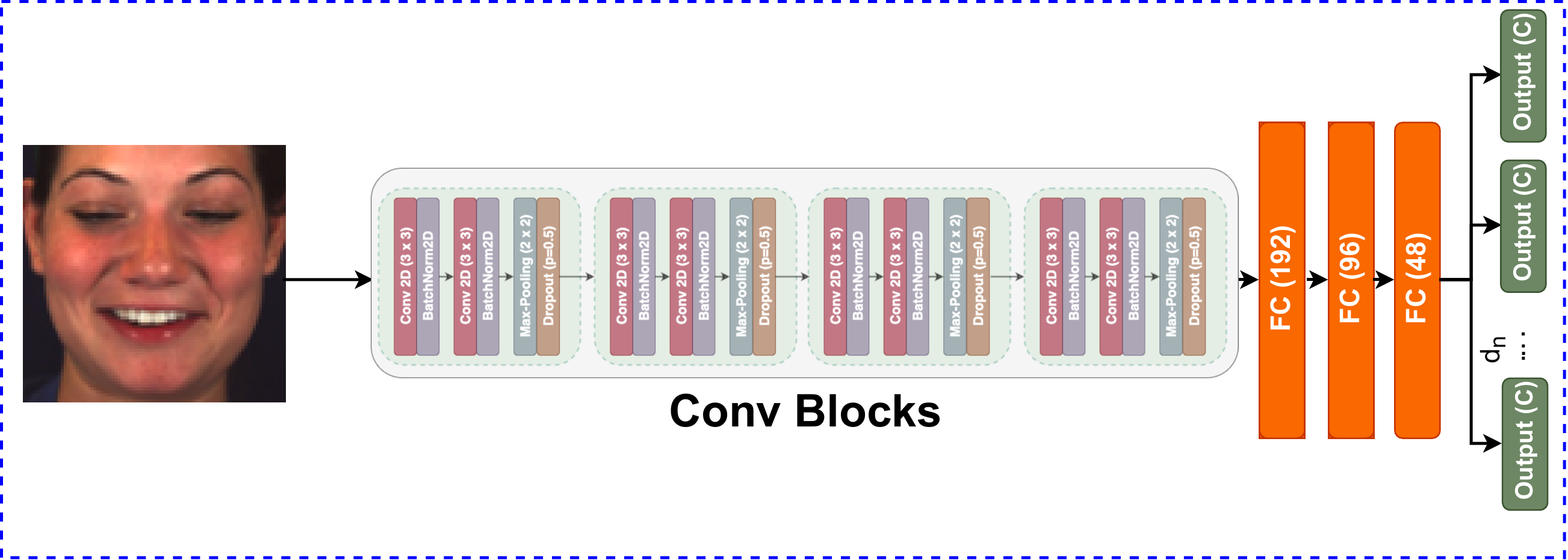}}\hspace{1em}
    \subfloat[\acs{DA} Architecture~\cite{xu2020investigating} separating task-specific and domain-specific features.\label{fig:disentangled}]{\includegraphics[width=0.267\textwidth,valign=c]{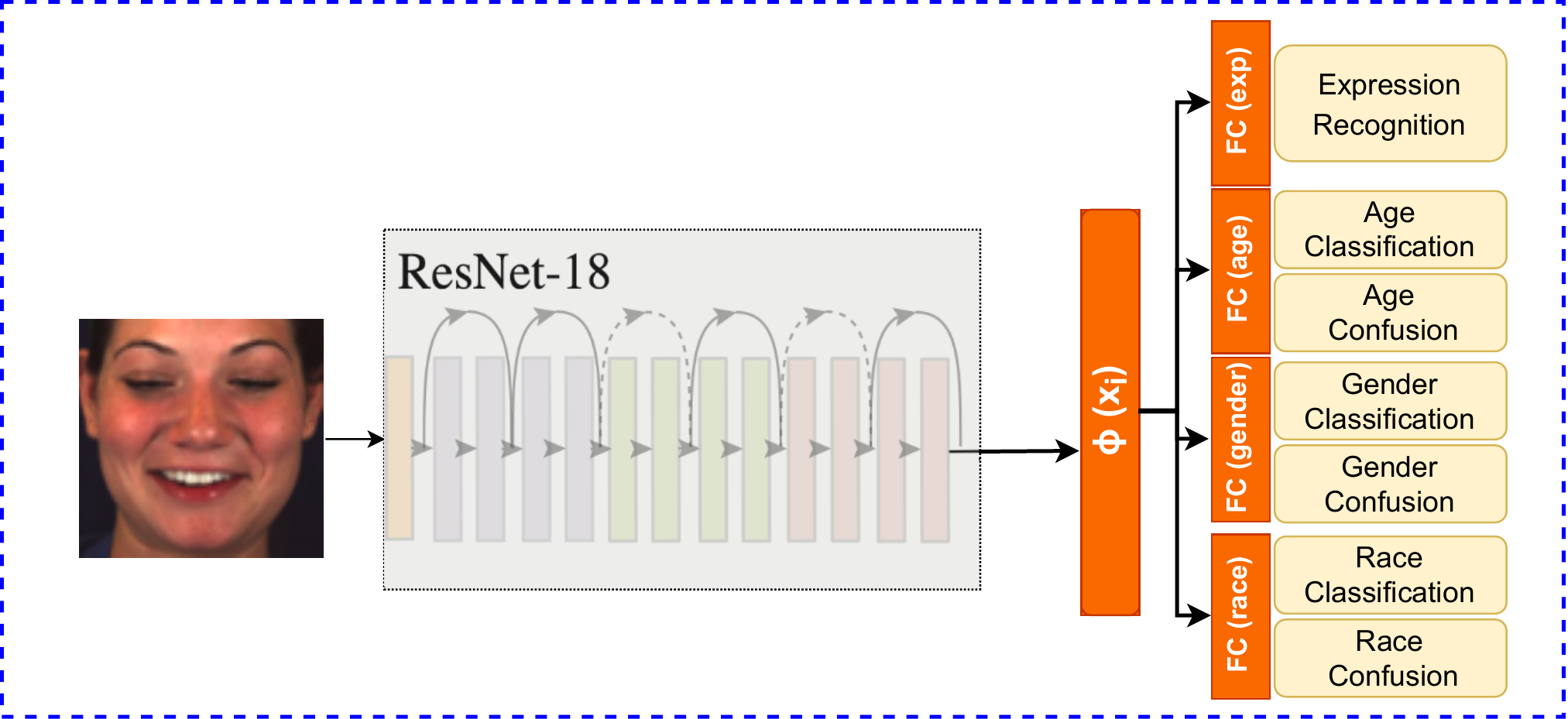}}
    % 
    % \subfloat{\includegraphics[width=0.58\textwidth]{Figures/Architectures/archs.pdf}}\hspace{0.2em}
    % \subfloat{\includegraphics[width=0.4\textwidth]{Figures/Architectures/da2.pdf}}
    \vspace{-2mm}\caption{Architectures for (a)~the Baseline CNN,~(b)~the \acf{DDC}~\cite{wang2020towards},~(c)~the \acf{DIC}~\cite{wang2020towards}, and~(d)~the \acf{DA}~\cite{xu2020investigating}.}
    \label{fig:model_architectures}
    % \vspace{2mm}
\end{figure*}

% \begin{figure*}
%     \centering
%     \includegraphics[width=0.7\textwidth]{Figures/arch.pdf}
%     \caption{Model Architectures for (a)~the Baseline CNN,~(b)~the \acf{DDC}~\cite{wang2020towards},~(c)~the \acf{DIC}~\cite{wang2020towards}, and~(d)~the \acf{DA}~\cite{xu2020investigating}.}
%     \label{fig:model_architectures}
% \end{figure*}
\section{Introduction}
% Facial expression recognition (FER) is a popular area of research in the fields of affective computing and computer vision due to its potential in human-computer interface and its role in human's social life. Owing to the development of artificial intelligence and machine learning, the fairness between demographic groups such as race, gender, age, etc. comes into prominence. Nevertheless, traditional machine learning algorithms have the risk of amplifying societal stereotypes. Moreover, mostly studied FER datasets do not always contain an equally distributed samples regarding their attributes. Thus, it is critial to ensure that these models are not affected by the inequalities in the dataset to avoid making biased decisions.

From security and surveillance systems~\cite{Feldstein2019AISurveillance}, monitoring emotional and mental wellbeing~\cite{Bodala2020creating} and assisting in medical interventions~\cite{Borenstein2017Pediatric} to law enforcement~\cite{Howard2017}, robots are becoming closely embedded in our society, making `smart' decisions about several critical aspects of our lives~\cite{Roselli2019}. Therefore, it is crucial that they make fair and unbiased decisions~\cite{Howard2017} in order to avoid potentially catastrophic consequences that adversely affect individuals~\cite{goodman2017european}. 

Fair and unbiased analysis and interpretation of human affective behaviours are among the factors that can contribute to the realisation of effective long-term \ac{HRI}. Successful long-term HRI can be used to provide physical and emotional support to the users, engaging them in a variety of application domains that require personalised human–robot interaction, including healthcare, education and entertainment~\cite{Churamani2020CL4AR}. 

\acf{FER} algorithms (see~\cite{Sariyanidi2015Automatic,nott44740,Li12020Deep} for a detailed survey) aim to analyse human facial expressions either by encoding facial muscle activity as Facial \ac{AU}~\cite{ekman1978facial} or determining the emotional state being expressed by an individual~\cite{Ekman1971Constants, Ekman2009}. Analysing large datasets of human faces, annotated for the expressions represented in the images, training \ac{FER} models becomes heavily data-dependent and may be prone to \textit{biases} originating from imbalances in the data distribution with respect to attributes like gender, race, age or skin-color, implicitly encoded in the data~\cite{Li2020Deeper}. These imbalances may result in the models learning to associate such \textit{confounding} attributes with the task of \ac{FER}. With affective robots increasingly becoming an integral part of daily human life, `bias' in \ac{FER} models, as described above, may result in unfavourable and prejudiced consequences for many under-represented groups. 

Most recent and popular \ac{FER} datasets provide annotations for different demographic attributes, along with affective labels, allowing for \textit{fairer} \ac{FER} models using these annotations explicitly~\cite{Li2020Deeper}. This can be done at the \textit{pre-processing} level~\cite{yucer2020exploring} either modifying the data distribution in favour of underrepresented groups while training by strategically sampling~\cite{elkan2001foundations} the data, or at \textit{in-processing} level~\cite{yucer2020exploring} by adapting the model architecture or the training process to handle these imbalances. Some methods achieve this by  either forcing the model to explicitly learn domain-specific information such that this can be discounted from the model's learning later~\cite{Dwork2012Fairness} or by discounting domain-specific information by omitting these features from the learnt representations~\cite{wang2020towards}. Other popular strategies include using data-augmentation strategies to synthetically generate additional data for the under-represented groups~\cite{Han2019Adversarial,Abbasnejad2017Using,CHARTE2015MLSMOTE, churamani2020clifer} to balance training data distribution or weighting model prediction loss differently for the different domain attributes. A weighting factor may be applied to the loss computation based on the occurrence rate for the different classes or domains~\cite{Shao2018Deep, Churamani2020AULACaps, elkan2001foundations} penalising misclassifications for under-represented groups more than others. The underpinning principle behind these methods, however, remains the same, focusing on balancing data distributions or learning to adapt to the imbalances by adjusting the learning algorithm.

\acf{CL} approaches~\cite{parisi2019continual, LESORT2020CL4R} focus on this very challenge of managing shifts in data distributions by \textit{continually} learning and adapting to novel information without forgetting previously learned information. As agents interact with their environments and gather more information about specific tasks, they need to be able to remember previously learnt tasks while acquiring new skills. \ac{CL} may allow them to balance learning across different domains or tasks as well as being robust against the changes in the data distributions. In particular, Domain-Incremental \ac{CL}~\cite{van2019three} deals with managing shifts in the input data distributions, while the task to be learnt remains the same. This can be considered analogous to managing affective interactions with users belonging to different domain groups of gender (male or female) or race (black, white or asian), trying to analyse their expressions.

% Despite the model learning the same task, the underlying data distribution changes as the model continues to adapt its learning to different domains.

We propose the use of \ac{CL} algorithms, benefiting from their ability for \textit{lifelong} learning, to develop \textit{fairer} \ac{FER} models for affective robots that can balance learning with respect to different attributes of \textit{gender} and \textit{race}. In particular, we formulate expression recognition and \ac{AU} detection, across these different domain attributes, as continual learning problems. We compare several popular regularisation-based \ac{CL} approaches with state-of-the-art bias mitigation methods by splitting the data into different domain labels for each attribute; \textit{male} and \textit{female} for gender, \textit{White/Caucasian, Black/African-American, Asian} and \textit{Latino} for race, provided by the RAF-DB and BP4D datasets. \ac{CL}-based approaches are, on average, seen to outperform other bias mitigation strategies both in terms of accuracy as well as fairness for both the RAF-DB dataset (across gender and race splits) and the BP4D dataset (for race splits).

% Exploring Domain-Incremental \ac{CL} settings for learning facial expression analysis tasks, split along demographic attributes of gender and race, the models need to learn to solve these tasks, one domain-group at a time, robustly dealing with the changes in the data distribution across these groups. 

% The contributions of this paper are as follows:\\
% (1) To the best of our knowledge, there is no study that systematically focusing on the bias problem across different domains in facial expression recognition except for \cite{xu2020investigating, wang2020towards}.
% (2) Different from previous studies, our work makes a first step to use continual learning approaches as a new strategy for building fairer models.
% (3) Extensive experiments have been conducted on two well-established facial expression datasets: RAF-DB and BP4D. Comparing with other methods, continual learning approaches outperform Non-CL approaches in terms of fairness.

\section{Methodology}
\label{sec:approaches}
To investigate the problem of bias in \ac{FER} systems, we need to understand which domain attributes dominate the data and how an algorithm performs with respect to these attributes. In this section, we present the problem formulation, the learning scenario as well as briefly describe the different non-\ac{CL} and \ac{CL}-based methods compared in this work.

\subsection{Problem Formation}
As our objective is to evaluate bias in \ac{FER} with respect to different domain attributes, we focus on evaluating model performances on datasets split across \textit{gender} and \textit{race} domain attributes on two different facial analysis tasks: Expression recognition and \ac{AU} detection. Thus, given a set of samples $x_i$, ground truth labels $y_i$ and domain labels $d_i$ we are evaluate the performance of the model $\mathcal{A}(x_i|y_i, d_i)$ across the different domain labels.

We implement a ResNet-based \ac{CNN}~\cite{He2016resnet} architecture composed of $4$ conv blocks each of which consists of $2$ convolutional layers, a max pooling layer, with dropout and batch normalisation. The output of the last conv block is attached to a three-layered \ac{MLP} with a classifier output. ReLU activation is used after all conv and dense layers. We use the same \acs{CNN} architecture as the basis for all the approaches, except for the \acf{DA}~\cite{xu2020investigating} where we take the results from the original paper. 

\subsubsection{Baseline Model}
As our baseline approach, above described \acs{CNN} model (see Fig.~\ref{fig:model_architectures}) is incrementally trained on the domain-based splits of the dataset and model performance is reported after the training for each split. This method is commonly referred to as \textit{finetuning}~\cite{aljundi2018memory}.

\subsubsection{Offline Training} 
As a second baseline, we train the above-described \textit{baseline} \acs{CNN} model with all the training data, \textit{off-line}, at once but report its performance scores individually on domain-specific test-splits. 
	
% \end{enumerate}

\subsection{Non-\ac{CL}-based Bias Mitigation Strategies}
In this section, we describe four popular and state-of-the-art bias mitigation methods, grouped under `non-\ac{CL}-based' approaches, to be compared with \ac{CL}-based solutions. 

\subsubsection{\textbf{\acf{DDC}}:} A popular approach for bias mitigation in \acs{ML} systems is referred to as ``\textit{fairness-through-awareness}''~\cite{Dwork2012Fairness} where the domain information is explicitly learnt and encoded in feature representations making the model more `aware' of the different domain labels in order to discriminate between each of them. The model implements a $N\times M$-way classifier~\cite{wang2020towards} where $N$ is the number of domains and $M$ is the number of classes to be learned per domain (see Fig~\ref{fig:model_architectures}).

\subsubsection{\textbf{\acf{DIC}}:} One of the concerns with \ac{DDC}-based models is that they may learn decision boundaries for the same class across different domains, that is, even if the model predicts the correct class, it may be penalised unnecessarily due to differences in domain-specific features. \et{Wang}~\cite{wang2020towards} propose a different approach by training separate classifiers for each of the domain while sharing the same top-level architectures. This, with the different model heads (see Fig~\ref{fig:model_architectures}), it can learn to solve the task for each domain group individually and independently.

\subsubsection{\textbf{\acf{SS}}:} The simplest approach to mitigate bias arising from skewed data distributions, without changing model architecture, is to balance the effect of each domain distribution by \textit{strategically sample} data~\cite{elkan2001foundations} for each domain-class mapping. This can be achieved by increasing the sampling frequency of the images from under-represented distributions or equivalently, for each domain $d_i$ assigning a loss weight $w_i$ inversely proportional to the rate of occurrence of sample for that domain. We follow the weighted-loss approach for strategically sampling data.

\subsubsection{\textbf{The \acf{DA}}:} \et{Xu}~\cite{xu2020investigating} implement the \acf{DA}~\cite{Liu_2018_CVPR} that aims to mitigate bias across sensitive domains by making sure that the feature representations learnt by the model do not contain any domain-specific information. The network is split into two parts with a shared ResNet-based feature extraction sub-network. The first part focuses on facial affect analysis, while the other part consists of separate branches for each domain, designed to suppress domain-specific information. For our experiments, the results for the \ac{FER} tasks are taken from the original paper~\cite{xu2020investigating}.
	
\subsection{Continual Learning Approaches}
In this work, we primarily explore regularisation-based \ac{CL} approaches, under \ac{Domain-IL} settings, as these can be implemented with the least additional computational and memory overhead. For a comparison with rehearsal-based methods, we also implement a simple \acf{NR} \cite{Hsu18_EvalCL} method that physically stores previously encountered data samples for rehearsal. Other \ac{CL}-based methods that improve model performance using a generative or probabilistic model to simulate \textit{pseudo-samples} for previously seen tasks~\cite{Robins1995, shin2017continual,churamani2020clifer} or by dynamically expanding model architectures by adding dedicated neurons (Growing Neural Networks~\cite{Parisi2018a,churamani2020clifer}) or neural layers (Progressive Networks~\cite{rusu2016progressive}) sensitive to specific domains or tasks are omitted from this evaluation as they require additional memory and computational resources to be allocated, giving them an unfair advantage over non-\ac{CL}-based bias mitigation methods. %This allows a fair comparison with the other bias-mitigation approaches discussed in the paper. 
For our experiments, the models need to learn \ac{AU} detection and \ac{FER} tasks as the input data distributions change with respect to domain-specific attributes of \textit{gender} and \textit{race}. We use the same baseline \ac{CNN} model and apply the learning protocol as described by the following approaches:

% We specifically omit approaches focusing on dynamically adapting and expanding model architectures, that is, adding dedicated neurons or additional neural layers sensitive to specific domains or tasks, as in the case of Progressive Networks~\cite{rusu2016progressive} and growing neural networks~\cite{Parisi2018a,churamani2020clifer} despite their success on \ac{CL} tasks, to allow for a fair comparison with the other approaches discussed in the paper with constrained neural resources.

\subsubsection{\textbf{\acf{EWC}}:} As proposed by \et{Kirkpatrick}~\cite{kirkpatrick2017overcoming}, the \ac{EWC} approach introduces a quadratic penalty determined by the relevance of each parameter of the model with respect to old and new tasks, penalising updates in parameters relevant for old tasks, in order to avoid forgetting previously learnt information. The importance of different model parameters are determined using a Fisher Information Matrix, updating the loss function.

\subsubsection{\textbf{\ac{EWC}-Online}:} A disadvantage for the \ac{EWC} method is that, as the number of tasks increase, the number of quadratic terms for regularisation grows. To handle this, \et{Schwarz}~\cite{schwarz2018progress} proposed a modification to \ac{EWC} where instead of many quadratic terms, a single quadratic penalty is applied, determined by a running sum of the Fischer Information Matrices of the previous tasks.

\subsubsection{\textbf{\acf{SI}}:} Similar to \ac{EWC}, the \ac{SI} approach also penalises changes to relevant weight parameters (synapses) such that new tasks can be learnt without forgetting the old~\cite{zenke2017continual}. To avoid forgetting, importance for solving a learned task is computed for each parameter and changes in most important parameters are discouraged.

\subsubsection{\textbf{\acf{MAS}}:} The MAS approach also tries to alleviate forgetting by calculating the importance of each parameter by looking at the sensitivity of the output function instead of the loss~\cite{aljundi2018memory}. Parameters that have the most impact on model predictions are given a high importance and changes to these parameters are penalised. However, unlike \ac{EWC} and \ac{SI}, parameter importance is calculated in an unsupervised manner with using only unlabelled data.
	
\subsubsection{\textbf{\acf{NR}}:} Inspired by the model used in~\cite{Hsu18_EvalCL}, we implement a (naive) rehearsal-based approach that implements a small replay-buffer to randomly store a fraction of previous seen data. This \textit{old data} along with the new data is used to create mini-batches using equal number of samples from both old and new data and used to train the model ensuring that old knowledge is not overwritten by new data.

\section{Experimentation and Results}

\subsection{Set-up}

For our experiments, we conduct two separate evaluations, comparing the different bias mitigation approaches (see Section~\ref{sec:approaches}) for facial expression recognition and \ac{AU} detection tasks with the RAF-DB~\cite{li2017reliable} and BP4D~\cite{ZHANG2014BP4D} datasets, respectively. Each approach is compared in terms of the \textit{Fairness Scores} achieved, both with and without data-augmentation. All evaluations are repeated $3$ times and results are averaged across the repetitions, except for \ac{DA} where the results are taking from the original paper. 

\subsection{Fairness Measure}
To compare the different approaches for their \textit{fairness} with respect to model performance (in this case, \textit{Accuracy}) for different attributes of gender and race, we use the `equal opportunity' definition of \textit{fairness}, as proposed by \et{Hardt}~\cite{hardt2016equality}.  

Let \textbf{x}, \textbf{y}, \textbf{ŷ} be the variables denoting input, ground truth label and the predicted label, respectively, $s\in S_i$ be the sensitive (domain) attribute (for example, $(S_i = \{$male, female$\})$,  $f$ be a function computing the \textit{accuracy score} for a given sensitive attribute $s$ and $d$ be the dominant attribute which has the highest accuracy score, then the \textit{Fairness Measure} $\mathcal{F}\in[0,1]$ of a model is defined as the \textit{largest accuracy gap} among all sensitive attributes computed as the minimum of the ratios of the accuracy scores of each sensitive attribute with respect to the dominant attribute.

{\small
\begin{equation}
    \mathcal{F}  = \min_{}(\frac{f(\mathbf{\hat{y}}, \mathbf{y}, s_0, \mathbf{x})}{f(\mathbf{\hat{y}}, \mathbf{y}, d, \mathbf{x})}, ..., \frac{f(\mathbf{\hat{y}}, \mathbf{y}, s_n, \mathbf{x})}{f(\mathbf{\hat{y}}, \mathbf{y}, d, \mathbf{x})})\\
\end{equation}
}

\begin{table}[t]
	\centering
	\caption{\textbf{Experiment 1:} Fairness Scores across Gender and Race for the RAF-DB Dataset. \textbf{Bold} values denote best while [\textit{bracketed}] denote second-best values for each column.}
	\label{tab:rafdb_fair}
	\vspace{-2mm}
	{
	\footnotesize

	\begin{tabular}{c|c|c|c|c}\toprule
		\multirow{2}{*}{\textbf{Method}} & \multicolumn{2}{c|}{\textbf{W/O Data-Augmentation}} & \multicolumn{2}{c}{\textbf{W/ Data-Augmentation}} \\ \cmidrule{2-5} 
		
		 & \textit{\textbf{Gender}} & \textit{\textbf{Race}} %& \textit{\textbf{Age}} 
		 & \textit{\textbf{Gender}} & \textit{\textbf{Race}} %& \textit{\textbf{Age}} 
		 \\ \midrule
		% \rowcolor{gray!10}
		Baseline & 0.834 & 0.943 %& 0.867 
		& 0.816 & 0.937  %& 0.847 
		\\ 
		
% 	 	\rowcolor{gray!10}
		Offline & 0.944 & 0.925 %& 0.941 
		& 0.954 & 0.974 %& 0.932 
		\\ \midrule

		\multicolumn{5}{c}{\textbf{Non-\ac{CL}-based Bias Mitigation Methods}} \\ \midrule

		% \rowcolor{gray!10}
		\ac{DDC}~\cite{wang2020towards} & 0.968 &  0.985  %& 0.922 
		& 0.961 & 0.976  %& 0.926 
		\\ 

% 		 \rowcolor{gray!10}
		\ac{DIC}~\cite{wang2020towards} & 0.938 & 0.989 %& 0.955 
		& 0.962 & 0.965 %& 0.927 
		\\ 
		% \rowcolor{gray!10}
		\ac{SS}~\cite{elkan2001foundations} & 0.955 & 0.961 %& 0.941 
		& 0.954 & 0.975 %& 0.930 
		\\ 
% 		\rowcolor{gray!10}

		\ac{DA}~\cite{xu2020investigating} & 0.975 & 0.858 %& 0.903 
		& [\textit{0.997}] & 0.919 %& 0.889 
		\\ \midrule

		\multicolumn{5}{c}{\textbf{Continual Learning Methods}} \\ \midrule
		% \rowcolor{gray!10}
		\ac{EWC}~\cite{kirkpatrick2017overcoming} & 0.972 & 0.987 %& 0.952 
		& 0.983 & 0.990 %& 0.952 
		\\ 
% 		 \rowcolor{gray!10}
		\ac{EWC}-Online~\cite{schwarz2018progress} & 0.970 & 0.987 %& 0.964 
		& 0.974 & 0.990 %& 0.970 
		\\ 
		% \rowcolor{gray!10}
		\ac{SI}~\cite{zenke2017continual} & \cellcolor{gray!25}\textbf{0.990} & \cellcolor{gray!25}\textbf{0.996} %& \textbf{0.992} 
		& \cellcolor{gray!25}\textbf{0.999} & \cellcolor{gray!25}\textbf{0.996} %& \textbf{0.977} 
		\\ 
% 		 \rowcolor{gray!10}
		\ac{MAS}~\cite{aljundi2018memory} & [\textit{0.980}] & [\textit{0.990}] %& [\textit{0.982}] 
		& 0.990 & [\textit{0.994}] %& [\textit{0.971}] 
		\\ 
		% \rowcolor{gray!10}
		\acs{NR}~\cite{Hsu18_EvalCL} &0.928 & 0.974 %& 0.941 
		& 0.923 & 0.974 %& 0.946 
		\\ \bottomrule
\end{tabular}}
\vspace{-3mm}
\end{table}

\subsection{Experiment 1: Facial Expression Recognition}
To investigate the applicability of \ac{CL}-based methods as \textit{`fair'} \ac{FER} systems, we train and test the approaches described in Section~\ref{sec:approaches} on the RAF-DB dataset and compare their performance (both without and with data augmentation) on learning to correctly categorise the $7$ expression classes, namely, \textit{surprise, sadness, happiness, fear, anger, disgust} and \textit{neutral}, with respect to $2$ different domain groups; \textit{gender} (Male, Female) and \textit{race} (Caucasian, African-American, Asian). The different approaches are trained with samples belonging to one domain-split at a time, actively trying to preserve the knowledge from previously seen splits while learning to classify samples for the new domain-split. As a result, \ac{CL} approaches, on average, outperform all other methods on their Fairness Scores with respect to both gender and race domains with SI yielding the best results both with and without data-augmentation (see Table~\ref{tab:rafdb_fair}). In addition, although data-augmentation has an overall positive effect on model accuracy for all approaches, no significant shift is witnessed in model fairness scores.

Furthermore, for the non-CL-based methods, except for \ac{DA}, complete knowledge of the domain-groupings is required apriori in order to design the architectures of the models, limiting their real-world applicability. For \ac{CL}-based methods, however, learning can be adapted to new domains \textit{on-the-fly} as the models are designed to be sensitive to random and sudden changes in data distributions encountering samples from different domain groups over time.

% Note that, we also do experiments to analyze the effect of the class-ordering, which refers to model performance being sensitive to the changes in the order of learning~\cite{churamani2020clifer}. However, for different arrangements in the order of gender and race groups, no significant effect of domain ordering is observed.   

\subsection{Experiment 2: Action Unit Detection}

\acf{AU} detection poses a multi-label classification problem where the models need to predict multiple \acsp{AU} activated in a given sample. As in the case of Experiment~$1$, we report and compare the different bias \ac{CL}-based and non-\ac{CL}-based methods on the BP4D dataset with respect to their average Fairness Scores across $12$ \acsp{AU}, individually for gender and race domain-splits. We see that even though the \ac{CL}-based methods are able to achieve highest individual accuracy scores for most of the gender and race groups, this comes at the cost of balancing learning across the different attributes. For the gender- splits, the \acf{DA}~\cite{xu2020investigating} achieves the highest fairness scores, despite performing moderately in terms of accuracy on individual splits (see Table~\ref{tab:bp4d_fairness}). In the case of race splits, we see that even though the \ac{NR} approach achieves the highest fairness cores, this is owed to the memory-intensive rehearsal mechanism that physically stores and replays samples from previously seen domains to retain model performance. Even though the regularisation-based approaches target accuracy and trade-off fairness in the process, they still perform better than most non-\ac{CL}-based methods. On the contrary, for the non-\ac{CL} methods we see that more importance is given to fairness than individual accuracy with \ac{DA} achieving consistently high fairness scores.

\begin{table}[t]
	\centering	
	\
	{
	\footnotesize
	\caption{\textbf{Experiment 2:} Fairness Scores across Gender and Race for the BP4D Dataset. \textbf{Bold} values denote best while [\textit{bracketed}] denote second-best values for each column.}
	\label{tab:bp4d_fairness}
	\vspace{-2mm}

	\begin{tabular}{c|c|c|c|c}\toprule
	
		\multirow{2}{*}{\textbf{Method}} & \multicolumn{2}{c|}{\textbf{W/O Data-Augmentation}} & \multicolumn{2}{c}{\textbf{W/ Data-Augmentation}} \\ \cmidrule{2-5} 
		
		 & \textit{\textbf{Gender}} & \textit{\textbf{Race}} & \textit{\textbf{Gender}} & \textit{\textbf{Race}} \\ \midrule
		 
		Baseline & 0.962 & 0.855 & 0.941 & 0.858 \\ 
		
		Offline & 0.984 & 0.878 & [\textit{0.994}] & 0.901 \\ 
		\midrule
		\multicolumn{5}{c}{\textbf{Non-\ac{CL}-based Bias Mitigation Approaches}} \\ \midrule
		
		\ac{DDC}~\cite{wang2020towards} & [\textit{0.990}] & 0.920 & 0.991 & 0.924 \\ 
		
		\ac{DIC}~\cite{wang2020towards} & 0.979 & 0.925 & 0.985 & 0.922 \\ 
		
		\ac{SS}~\cite{elkan2001foundations} & 0.977 & 0.920 & 0.983 & 0.919 \\ 
		
		\ac{DA}~\cite{xu2020investigating} & \cellcolor{gray!25}\textbf{0.994} & [\textit{0.954}] & \cellcolor{gray!25}\textbf{0.995} & [\textit{0.962}] \\ \midrule
		
		\multicolumn{5}{c}{\textbf{Continual Learning Approaches}} \\ \midrule
		
		\ac{EWC}~\cite{kirkpatrick2017overcoming} & 0.981 & 0.949 & 0.992 & 0.943 \\ 
		
		\ac{EWC}-Online~\cite{schwarz2018progress} & 0.976 & 0.937 & [\textit{0.994}] & 0.957 \\ 
		
		\ac{SI}~\cite{zenke2017continual} & 0.986 & 0.946 & 0.965 & 0.954 \\ 

		\ac{MAS}~\cite{aljundi2018memory} & 0.966 & 0.920 & 0.967 & 0.909 \\ 
		
		\ac{NR}~\cite{Hsu18_EvalCL} & 0.983 & \cellcolor{gray!25}\textbf{0.966} & 0.954 & \cellcolor{gray!25}\textbf{0.974} \\ \bottomrule
	\end{tabular}}
% 	\vspace{-1mm}
\end{table}

\section{Conclusion and Future Work}

In this work, we focused on the problem of bias in facial analysis tasks and proposed a novel application of continual learning as a bias mitigation strategy for \ac{FER} systems. We highlight how using \ac{CL} can help develop \textit{fairer} expression recognition and \ac{AU} detection algorithms with our experiments with popular benchmark datasets; RAF-DB for expression recognition and BP4D for \ac{AU} detection showcasing the superlative performance of \ac{CL} methods at handling imbalances in data distributions with respect to demographic attributes of \textit{gender} and \textit{race}. In comparison with state-of-the-art bias mitigation approaches, these methods are able to balance learning across different domains, not only achieving high accuracy scores but also maintaining \textit{fairness} across the different splits. Yet, in our experiments we primarily focus on regularisation-based \ac{CL} methods due their efficacy and economic implementation for real-world application. It will be interesting to contrast these methods to other resource-hungry yet, improved algorithms~\cite{lopez2017gradient,rebuffi2017icarl} that are able to better handle long-term retention of knowledge. Combining regularisation-based model adaptation with latent replay strategies~\cite{Pellegrini2019Latent} may prove helpful in implementing \textit{fairer} facial analysis systems for long-term \ac{HRI}.

% Our benchmark experiments present motivating results towards \textit{fairer} facial expression analysis models for affective robots, aiming to improve individual user experience during human-robot interactions. 
Furthermore, future work for us also entails conducting long-term \ac{HRI} studies, comparing \ac{CL} vs. non-\ac{CL}-based methods, by embedding these models onto a humanoid robot. Implementing long-term social interactions with under-represented population groups such as children~\cite{Lai2016}, ethnic and racial minorities~\cite{buolamwini2018gender} and the elderly~\cite{broekens2009assistive} can help evaluate how \ac{CL}-based \ac{FER} systems respond to users from different demographics.

\begin{acronym}
    \acro{AC}{Affective Computing}
    \acro{ACC}{Average Accuracy}
    \acro{AI}{Artificial Intelligence}
    \acro{AU}{Action Unit}
    \acro{AUC}{Area Under The Curve}
    \acro{BP4D}{Binghamton-Pittsburgh 3D Dynamic (4D) Spontaneous Facial Expression Database}
    \acro{BWT}{Backward Transfer}
    \acro{CF}{Catastrophic Forgetting}
    \acro{CHL}{Competitive Hebbian Learning}
    \acro{CL}{Continual Learning}
    \acro{Class-IL}{Class Incremental Learning}
    \acro{CLS}{Complementary Learning Systems}
    \acro{CNN}{Convolutional Neural Network}
    \acro{DA}{Disentangled Approach}
    \acro{DDC}{Domain Discriminative Classifier}
    \acro{DIC}{Domain Independent Classifier}
    \acro{Domain-IL}{Domain Incremental Learning}
    \acro{EWC}{Elastic Weight Consolidation}
    \acro{FACS}{Facial Action Coding System}
    \acro{FER}{Facial Expression Recognition}
    \acro{FWT}{Forward Transfer}
    \acro{GAN}{Generative Adversarial Network}
    \acro{GWR}{Growing When Required}
    \acro{HRI}{Human-Robot Interaction}
    \acro{MAS}{Memory Aware Synapses}
    \acro{ML}{Machine Learning}
    \acro{MLP}{Multi-layered Perceptron}
    \acro{NC}{New Concepts}
    \acro{NI}{New Instances}
    \acro{NIC}{New Instances and Concepts}
    \acro{NR}{Naive Rehearsal}
    \acro{RaaS}{Robotics as a Service}
    \acro{RAF-DB}{Real-world Affective Faces Database}
    \acro{RL}{Reinforcement Learning}
    \acro{ROC}{Receiver Operating Characteristics}
    \acro{SAR}{Socially Assistive Robotics}
    \acro{SI}{Synaptic Intelligence}
    \acro{SS}{Strategic Sampling}
    \acro{Task-IL}{Task Incremental Learning}
\end{acronym}
\newpage
% \balance
\bibliographystyle{ACM-Reference-Format}
\bibliography{main.bib}

\end{document}